\documentclass[10pt, a4paper, conference, compsocconf]{IEEEtran}
\usepackage{ifpdf}

\ifCLASSINFOpdf
 \usepackage[pdftex]{graphicx}
  \graphicspath{{./}}
  \DeclareGraphicsExtensions{.pdf,.jpeg,.jpg,.png}
\else
  \usepackage[dvips]{graphicx}
  \usepackage{epsfig}
  \graphicspath{{figures/}}
  \DeclareGraphicsExtensions{.eps}
\fi

\usepackage[cmex10]{amsmath}
\usepackage[caption=false,font=footnotesize]{subfig}
\usepackage{url}

\newif\ifdotikz\dotikzfalse
\newif\ifmakebbl

\usepackage{fancybox}
\usepackage{latexsym}
\usepackage{xspace}
\usepackage{amstext,amssymb,amsbsy}
\usepackage{alltt}             
\usepackage{epic,eepic}  
\usepackage{epstopdf}        
\usepackage{pifont}
\usepackage{color}
\usepackage{paralist}
\usepackage{enumerate}
\usepackage{float}
\usepackage{multirow}
\usepackage[lined,algoruled]{algorithm2e}
\ifdotikz
\usepackage{pgf}
\usepackage{tikz}
\usetikzlibrary{decorations.markings}
\usetikzlibrary{arrows,shapes,snakes,backgrounds,positioning}
\usetikzlibrary{fit}
\pgfrealjobname{paper}
\else
\ifpdf
\long\def\beginpgfgraphicnamed#1#2\endpgfgraphicnamed{\includegraphics[scale=0.5]{#1}}
\else
\long\def\beginpgfgraphicnamed#1#2\endpgfgraphicnamed{\epsfig{file=#1.eps}}
\fi
\fi

\newcommand{\nop}[1]{}

\newcommand{\norm}[1]{\left\lVert#1\right\rVert}
\newcommand{\nnorm}[1]{\norm{#1}_{2}}
\newcommand{\degree}{\ensuremath{^\circ}}

\newcommand{\danet}{DA-net}
\newcommand{\blnet}{BL-net}
\newcommand{\trt}[1]{\texttt{TRACK~#1}}
\newcommand{\spac}{\ensuremath{p_{\mathit{ld}}}}
\newcommand{\dblp}{\ensuremath{p_{\mathit{dp}}}}
\newcommand{\lnds}{\ensuremath{p_{\mathit{ls}}}}
\newcommand{\notxt}{\ensuremath{p_{\mathit{nt}}}}
\newcommand{\spact}{\emph{leading}}
\newcommand{\dblpt}{\emph{double-page}}
\newcommand{\lndst}{\emph{landscape}}
\newcommand{\notxtt}{\emph{noText}}

\begin{document}
\title{Baseline Detection in Historical Documents using Convolutional U-Nets}

\author{\IEEEauthorblockN{Michael Fink, Thomas Layer, Georg Mackenbrock, and Michael Sprinzl}
\IEEEauthorblockA{Deutsches Medizinrechenzentrum GmbH \& Co KG\\
Vienna, Austria\\
Email: \{fink,layer,mackenb,sprinzl\}@dmrz.de}
}

\maketitle

\begin{abstract}
Baseline detection is still a challenging task for heterogeneous collections of 
historical documents. 
We present a novel approach to baseline extraction in such settings, 
turning out the winning entry to the ICDAR 2017 Competition on Baseline detection (cBAD).  
It utilizes deep convolutional nets (CNNs) for both, the actual extraction of baselines,  
as well as for a simple form of layout analysis in a pre-processing step. To the best of our knowledge it is the first CNN-based system for baseline extraction applying a U-net architecture and sliding window detection, profiting from a high local accuracy of the candidate lines extracted. Final baseline post-processing complements our approach, compensating for inaccuracies mainly due to missing context information during sliding window detection. 
We experimentally evaluate the components of our system individually on the cBAD dataset. 
Moreover, we investigate how it generalizes to different data by means of
the dataset used for the baseline extraction task of the ICDAR 2017 Competition on Layout Analysis for Challenging Medieval Manuscripts (HisDoc). A comparison with the results reported for HisDoc shows that it also outperforms the contestants of the latter.
\end{abstract}


\IEEEpeerreviewmaketitle

%%%%%%%%%%%%%%%%%%%%%%%%%%%%%%%%%%%%%%%%%%%%%%%%%%%%%%%%%%%%%%%%%%%%%%%%%%%%%%%%%%%%
\section{Introduction}
\label{sec:introduction}

Textline segmentation is an essential preprocessing step for handwritten text recognition (HTR)~\cite{DBLP:conf/icdar/RomeroSBDD15},
and still a challenging task for heterogeneous collections of 
historical documents. In particular, if they are of an irregular and complex structure or of low quality such as faint typing. Moreover, ancient documents often exhibit further degradations such as, e.g., bleed-through, holes, ornamentation, annotations, etc.~(cf.~\cite{DBLP:journals/ijdar/Likforman-SulemZT07}, for instance).
Recent competitions, either on textline extraction itself~\cite{DBLP:journals/corr/Diem,DBLP:conf/icdar/MurdockRHR15} or including it as a subtask towards HTR~\cite{DBLP:journals/corr/Simistra,DBLP:conf/icfhr/SanchezRTV16}, aim at covering a wide range of these difficulties in order to challenge and assess state-of-the-art methods and algorithms.

We present a novel approach to baseline extraction in historical documents that utilizes deep convolutional nets (CNNs) and, to the  best of our knowledge, is the first to apply a U-net architecture~\cite{DBLP:journals/pami/ShelhamerLD17,DBLP:conf/cvpr/HeZRS16} with sliding window detection. While sliding windows over high resolution images has been a highly successful technique to perform object detection in images with CNNs, it has largely been considered unsuited for the detection of handwritten textlines, where a relatively large number of possibly overlapping objects of the same class need to be extracted. 

\begin{figure*}[h!]
\begin{center}
\includegraphics[scale=0.23]{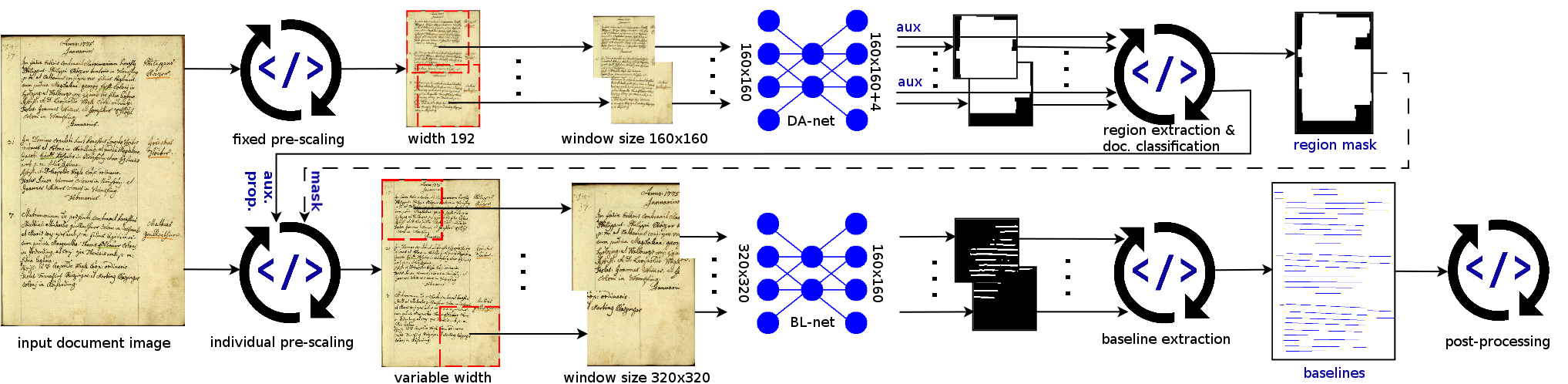}
\end{center}
\caption{Workflow of our approach 
including central components (\danet, \blnet, procedural units) and corresponding data flow.}
\label{fig:workflow}
\vspace*{-2ex}
\end{figure*}

We show that to the contrary, as far as the extraction of (poly-)baselines is concerned, a CNN-based sliding window approach can be very effective, witnessed for instance by  
winning the  ICDAR 2017 Competition on Baseline detection (cBAD)~\cite{DBLP:journals/corr/Diem}.
To profit from a high local accuracy of candidate baselines extracted by a CNN (in our case a residual U-net denoted as \blnet), however, our approach incorporates two further components: 
($i$) a second U-net (\danet), performing simple layout analysis wrt.~relevant text regions and auxiliary document properties, is applied in a pre-processing step and utilized, e.g., for individual document pre-scaling; ($ii$) final 
baseline post-processing aims at pruning spurious candidate lines and assembling baseline fragments into polygons, thus compensating for inaccuracies mainly due to missing context information during sliding window detection 
(cf. Fig.~\ref{fig:workflow} for an overview of the overall workflow). 
Quantifying the effect of the \danet~and post-processing on the overall task is one of the aspects covered by our experimental evaluation. Moreover, we investigate how it generalizes to different data by means of 
the dataset used for the baseline extraction task of the ICDAR 2017 Competition on Layout Analysis for Challenging Medieval Manuscripts (HisDoc).

\smallskip
Compared to most systems submitted to recent competitions on baseline extraction   
our approach likewise builds  
on neural net based machine learning techniques. 
However, 
those mainly apply recursive architectures (RNNs) (often using LSTM nodes, e.g.,  multi-dimensional LSTMS~\cite{DBLP:conf/icann/GravesFS07}) due to their capabilities to learn keeping track of context information (see, however,~\cite{DBLP:journals/pami/ShiBY17} for a mixed CNN and RNN architecture  
for baseline segmentation applied by 
a contestant of~\cite{DBLP:conf/icfhr/SanchezRTV16}). 
A notable exception from scene text recognition is~\cite{DBLP:conf/icpr/WangWCN12}, where a successful approach towards end-to-end printed text recognition in natural images has been developed by means of a CNN using sliding windows to detect a set of candidate lines of text. Textline segmentation was not the final aim but provided input to a second stage to obtain end-to-end results. Despite its success on printed text recognition, it was deemed unsuited for handwritten text.

More recently,~\cite{DBLP:conf/das/Pastor-Pellicer16}
used CNNs for textline extraction also aiming at a robust method explicitly focusing on historical documents. 
However, textlines are segmented at a different level of detail, returning a surrounding polygon (Main Body Area) at pixel level. Compared to poly-baselines, this usually requires considerable additional human labeling, resp.~correction, effort yielding only slightly higher HTR precision~\cite{DBLP:conf/icdar/RomeroSBDD15}.

In~\cite{DBLP:conf/icfhr/MoyssetLKW16} the challenge of handling high resolution images with a large amount of small objects is addressed using multiple local prediction models with shared parameters. Textline bounding boxes are obtained 
detecting lower-left and upper-right corner points in a first stage and matching point pairs in a second stage through local processing. 
The latter uses small spatial support requiring less resources (parameters). Besides requiring bounding boxes also for training, compared to our approach 
the maximum number of lines per image is fixed by 
the architecture and, moreover, 2-dimensional LSTM neurons are added to the local network to include context.

The remainder of this paper is organized as follows. 
Section~\ref{sec:sda} is devoted to  
simple document analysis, introducing the common overall U-net architecture of the CNNs used, as well as the concrete specification of the \danet. Actual baseline detection by means of the \blnet~is subject to Section~\ref{sec:bld}, while baseline post-processing is dealt with in Section~\ref{sec:bpp}. 
In Section~\ref{sec:exp} we evaluate the individual components of our approach on the cBAD dataset and compare with the results reported on the HisDoc competition, before we conclude in Section~\ref{sec:concl}.

%%%%%%%%%%%%%%%%%%%%%%%%%%%%%%%%%%%%%%%%%%%%%%%%%%%%%%%%%%%%%%%%%%%%%%%%%%%%%%%%%%%%
\section{Simple Document Analysis}
\label{sec:sda}
%%%%%%%%%%%%%%%%%%%%%%%%%%%%%%%%%%%%%%%%%%%%%%%%%%%%%%%%%%%%%%%%%%%%%%%%%%%%%%%%%%%%

The CNNs we apply 
follow a common  
U-net architecture~\cite{DBLP:journals/pami/ShelhamerLD17} 
(see Fig.~\ref{unet}), consisting of a contracting path (left path downward) and an expanding path (right path upward). 
Downscaling via the contraction path 
is realized using max-pooling layers after each of several blocks of layers. Corresponding upscaling is achieved using up-sampling via nearest-neighbor interpolation.  
Horizontal red arrows indicate the usage of equal resolution feature maps
from the contracting path as additional input for the expanding path, intended to 
enhance feature localization. 
In the following, we provide respective details 
for the \danet~architecture and describe its application to classify relevant text regions and auxiliary document properties.  

\ifdotikz
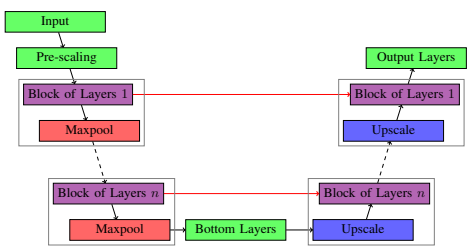
\begin{figure*}[h!]

\begin{center}
\beginpgfgraphicnamed{U-net-template}
\begin{tikzpicture}
  \tikzstyle{R}=[rectangle,draw=black,fill=red!60,text=black,minimum width=75pt,minimum height=15pt] 
  \tikzstyle{G}=[rectangle,draw=black,fill=green!60,text=black,minimum width=75pt,minimum height=15pt]
  \tikzstyle{B}=[rectangle,draw=black,fill=blue!60,text=black,minimum width=75pt,minimum height=15pt]
  \tikzstyle{V}=[rectangle,draw=black,fill=violet!60,text=black,minimum width=75pt,minimum height=15pt]

     %%% NODES from top left to top right
     \node[G](input){\text{Input}};
     \node[G,below=0.4cm of input,xshift=0.3cm](prescale){\text{Pre-scaling}};
     \node[V,below=0.4cm of prescale,xshift=0.3cm](bb1down){\text{Block of Layers $1$}};
     \node[R,below=0.4cm of bb1down,xshift=0.3cm](maxpool1){\text{Maxpool}};
     \node[V,below=1.1cm of maxpool1,xshift=0.5cm](bbndown){\text{Block of Layers $n$}}; 
     \node[R,below=0.4cm of bbndown,xshift=0.3cm](maxpooln){\text{Maxpool}};
     \node[G,right=0.4cm of maxpooln](bottombb){\text{Bottom Layers}};
     \node[B,right=0.4cm of bottombb,xshift=0.3cm](upscalen){\text{Upscale}};
     \node[V,above=0.4cm of upscalen,xshift=0.3cm](bbnup){\text{Block of Layers $n$}};
     \node[B,above=1.1cm of bbnup,xshift=0.5cm](upscale1){\text{Upscale}}; 
     \node[V,above=0.4cm of upscale1,xshift=0.3cm](bb1up){\text{Block of Layers  $1$}};
     \node[G,above=0.4cm of bb1up,xshift=0.3cm](bbout){\text{Output Layers}};

     %%%EDGES unet-path
     \draw[->](input) -- (prescale); 
     \draw[->](prescale) -- (bb1down); 
     \draw[->](bb1down) -- (maxpool1);
     \draw[->](maxpool1) -- (bbndown)[dashed]; 
     \draw[->](bbndown) -- (maxpooln);
     \draw[->](maxpooln) -- (bottombb);
     \draw[->](bottombb) -- (upscalen);
     \draw[->](upscalen) -- (bbnup);
     \draw[->](bbnup) -- (upscale1)[dashed]; 
     \draw[->](upscale1) -- (bb1up);
     \draw[->](bb1up) -- (bbout);

     %%%EDGES direct connections
     \draw[->,draw=red](bbndown) -- (bbnup); 
     \draw[->,draw=red](bb1down) -- (bb1up); 

     \node [draw=gray, fit= (bb1down) (maxpool1) ] {};
     \node [draw=gray, fit= (bbndown) (maxpooln) ] {};
     \node [draw=gray, fit= (bb1up) (upscale1) ] {};
     \node [draw=gray, fit= (bbnup) (upscalen) ] {};

\end{tikzpicture}
\endpgfgraphicnamed
\end{center}
\caption{General U-net architecture of \danet~and \blnet.}
\label{unet}
\end{figure*}
\else
\begin{figure}[ht]
\begin{center}
\includegraphics[scale=0.5]{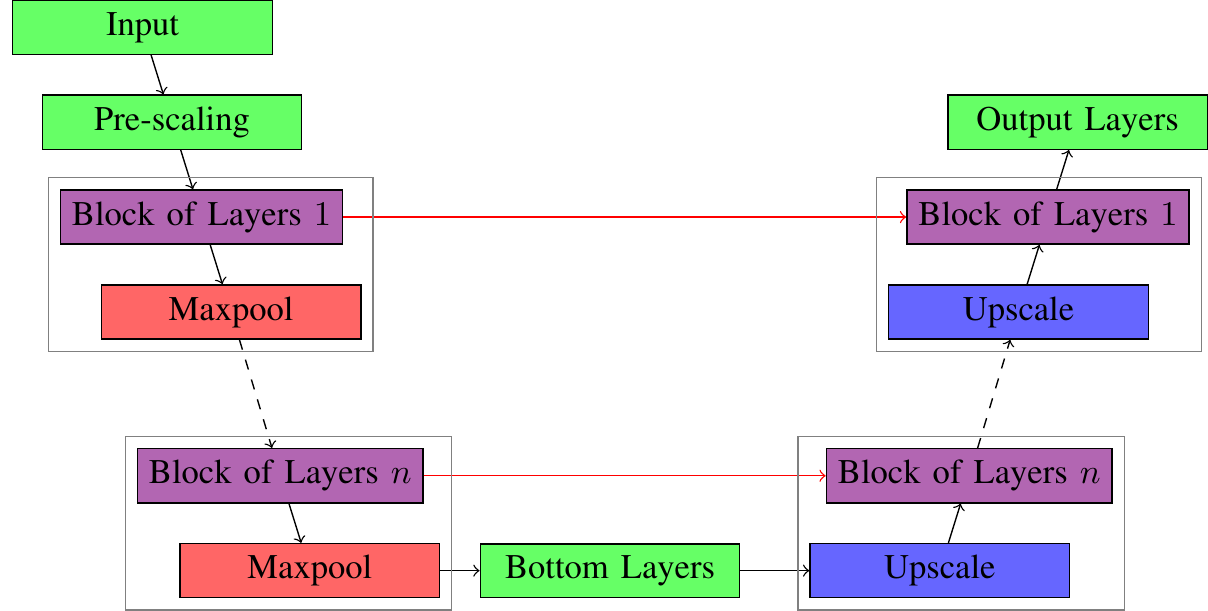}
\end{center}
\caption{General U-net architecture of \danet~and \blnet.}
\label{unet}
\vspace*{-2ex}
\end{figure}
\fi

\subsection{\danet~Architecture}\label{u-net_architecture}

The \danet~takes as its input images of size $160\times160$ pixels (clippings from document images). No further pre-scaling layers are applied. 
The corresponding target output (groundtruth) is a binary mask of the same size, representing whether the corresponding pixel belongs to a relevant text-region or not. 
The contracting path as well as the expanding path consist of four blocks of layers. The first two blocks of layers are composed of four convolution layers with a filter size of  $3\times3$, rectified linear units (ReLU) for activation, and batch normalization. 
Blocks 3 and~4 follow the same architecture, however, using two convolution layers. Max-pooling layers are applied with $2\times2$ filters and stride $2$, thus yielding a downscale by a factor of two. Upscaling doubles the size through bilinear up-sampling, effectively reversing this effect ($160\times160$ output nodes). 
Conversely, the number of channels is doubled after each block on the contracting path and halved after each block on the expanding path, starting with $32$ channels for the first block.
Hence, for the 
bottom layers we obtain $10\times10$ feature maps for 512 channels.

The bottom layers of the \danet~on the one hand connect contracting and expanding path as usual, in our case by means of two convolution layers with batchnorm as 
above. 
On the other hand we extended the 
U\nobreakdash-net architecture at this point, 
adding a fully connected layer with $512$ nodes and logistic activation 
after the last max-pooling layer. This layer is in turn connected to four 2-way softmax layers with corresponding auxiliary error layers (log probability). Intuitively, these layers are used 
for the classification of auxiliary document properties on the features of the contraction path (four additional outputs).  
  
The block of output layers connected to the expansion path again comprises two convolution layers as above, where the first one reduces to a single 
channel 
followed by a batchnorm layer, while the second uses logistic activation with an associated dice-coefficient error layer.

\subsection{Region Extraction and Document Classification}

Inference is realized through a sliding window approach. 
The input  document image is scaled to a fixed width of~$192$ pixels and a window of size $160\times 160$ is applied with stride~$40$. Note that in this case window size is large wrt.~document image size with the intention to learn/predict more global features. Every pixel of a window yields a prediction. Since windows overlap, we thus get up to $16$ predictions for every pixel of the input image (four in every direction). If one 
is greater than $0.5$ 
the pixel is considered foreground (white), and background (black) otherwise. 
A region mask is extracted 
restricting to 
connected components 
of at least ten foreground pixels (4\nobreakdash-connected). Each region 
(connected component) 
in the mask is approximated and represented through a polygon by vertically sampling the coordinates of leftmost and rightmost foreground pixels 
at a rate of $1/30$ image height.

Moreover, the auxiliary two-way softmax bottom layers provide four additional classification results, i.e., values from $[0,1]$ corresponding to class probabilities for the following document properties:
\begin{itemize}
\item \spact\ ($\spac$): 
the distance between successive baselines is large,
\item \dblpt\ ($\dblp$): 
the document spans two pages,
\item \lndst\ ($\lnds$): 
the orientation is landscape,
\item \notxtt\ ($\notxt$): 
the document does not contain text.
\end{itemize}
These predictions of document properties are used during 
baseline detection and post-processing, e.g., 
for scaling.

%%%%%%%%%%%%%%%%%%%%%%%%%%%%%%%%%%%%%%%%%%%%%%%%%%%%%%%%%%%%%%%%%%%%%%%%%%%%%%%%%%%%
\section{Baseline Detection}
\label{sec:bld}
%%%%%%%%%%%%%%%%%%%%%%%%%%%%%%%%%%%%%%%%%%%%%%%%%%%%%%%%%%%%%%%%%%%%%%%%%%%%%%%%%%%%

Candidate baselines are detected by means of a different, so-called residual U-net~\cite{DBLP:conf/cvpr/HeZRS16}, the \blnet.
It takes $320\times320$ pixel input images.
A $5\times5$ convolution layer with stride $2$ and ReLU activation is applied for prescaling. While still predicting the central $160\times160$ pixels (number of output nodes),  
the intuition 
is that the net may make use of surrounding context 
upon downscaling via this layer.
Groundtruth thus is again a binary mask, this time representing whether corresponding pixels (central $160\times160$) belong to a baseline. 
Contracting path as well as expanding path consist of 
five blocks of layers, each composed of three residual blocks: 
two $3\times 3$ convolutional layers with ReLU activation and batch normalization, followed by an add layer that adds the input of the first convolution layer and the output of the last one and applies a logistic. 
In all other aspects, the architecture coincides with that of the \danet, however, without 
auxiliary layers at the bottom.

Inference is again realized by 
sliding a window with stride $40$, 
this time 
of size 
$320\times 320$. 
Outputs correspond to predictions for the central $160\times160$ pixels of a window. Thereof, only the central $80\times80$ pixels (inner mask, cf.~next subsection) are taken into consideration. Hence, we obtain up to four predictions for every pixel of the input document image. Like for the \danet, these are turned into a classification of every pixel into foreground or background. 
Further on, candidate baselines are extracted from connected components with at least $50$ foreground pixels by applying the linear least square method. Note that thus only 
straight line segments 
are extracted.

\subsection{Dice Error Modifications}

\begin{figure}[t!]
    \centering
    \subfloat[$0.20$/$0.25$/$0.20$]{
        \includegraphics[width=0.12\textwidth]{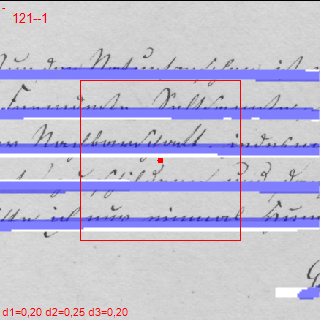}
        \label{fig:dice1}
    }
    ~ 
    \subfloat[$0.85$/$0.00$/$0.00$]{
        \includegraphics[width=0.12\textwidth]{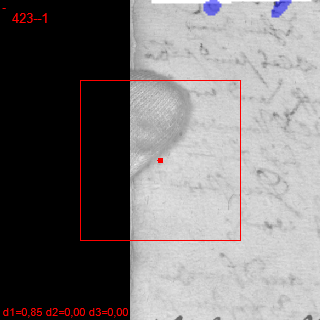}
        \label{fig:dice2}
    }
    ~ 
    \subfloat[$0.21$/$0.98$/$0.09$]{
        \includegraphics[width=0.12\textwidth]{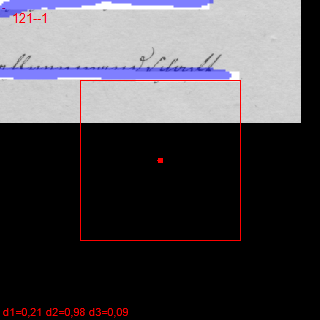}
        \label{fig:dice3}
    }
    \caption{Variants of the dice error coefficient on three samples (a-c): values for each sample correspond to the standard dice error/the dice error with inner mask (central square, equivalent to standard dice error on that square)/the dice error with inner mask and surrounding mask (3 pixels around inner mask, not drawn); blue regions represent prediction foreground, white regions represent groundtruth foreground.}\label{fig:dice}
\vspace*{-2ex}
\end{figure}

Since we employ a pixel-wise classifier trained wrt.~binary masks, 
we opted for an overlap metric for the error layer. In particular the dice coefficient is used which is a frequent choice to specify loss and assess the quality of segmentation (e.g., in medical image segmentation).  
However, we noticed 
some downsides with baselines near the border of a window: it is possible, that no text is visible but a groundtruth baseline reaches into the window (see Fig.~\ref{fig:dice2}) yielding a disproportional high error. For this reason we apply a mask such that just an inner region is evaluated. 
Of course, when the end 
of baseline just reaches into that inner region, again a relatively high error is obtained for masks that are close to the groundtruth (cf.~Fig.~\ref{fig:dice3}). To further mitigate this problem we use another region surrounding the inner region that is treated as correct, i.e., where we use groundtruth for evaluation.

More formally, consider $n\times n$ matrices $Y$ and $H$ denoting groundtruth and net 
output, respectively.
Given integers $a$ and $b$, such that $a\leq b\leq n$, we define a filter mask $M(a,b)$ as the $n\times n$ matrix given by  
$$
m_{ij}=\begin{cases}
1, & a\le i,j\le b \\
0, & else.
\end{cases}
$$
Let 
$M^I{=}\,M(a^I,b^I)$ be a filter mask (the \emph{inner mask}), 
let $M^Y{=}\,M(a^Y,b^Y)$ be another filter mask (\emph{surrounding mask}), such that $a^Y{<=}\,a^I$ and $b^Y{>=}\,b^I$, and let $\gamma\,{>}\,0$ be a constant.
Then, the modified dice coefficient  
is given by 
$$
D(H,Y)={\frac{2*|\bar Y\bar H|+\gamma}{|\bar Y|+ |\bar H|+\gamma}},
$$ 
where $\bar Y\,{=}\,YM^Y$, $\bar H\,{=}\,HM^I+Y(M^Y-M^I)$. 
Note that $a^Y{=}\,a^I$ and $b^Y{=}\,b^I$ yields the dice coefficient on the inner mask only, if moreover $a^Y{=}\,n$, then we obtain the dice coefficient on the full image. 

\subsection{Document Scaling}
\label{sec:DocumentScaling}

Another characteristic of our approach is the application of individual document scaling on the 
basis of document properties predicted by the \danet. To this end, we consider $n_{s}$ different scales and an upper bound  $w_{\mathit{max}}$ on document width (in our case $w_{\mathit{max}}\,{=}\,8000$ and $n_s\,{=}\,7$, namely 512, 640, 768, 896, 1024, 1152, and 1280 pixels). Given a document of width $w$ and height $h$, a scale index value $i_{s}$ is then calculated as follows:
\begin{equation*}
	\begin{aligned}
		i_{s} ={} & 
                \frac{n_{s} * w}{w_{\mathit{max}}} 
                + \frac{n_{s}}{4} * (2 + \lnds 
		           + \dblp - 4 * \spac).
	\end{aligned}
\end{equation*}
Intuitively, the first term represents an initial offset depending on the width of the document, 
while the second term may increase (landscape, double-page) or decrease this value (large leading).
If the document is (almost) empty, i.e., if $\notxt\,{>}\,0.7$, then this value is further 
decreased by $\frac{n_{s}}{4} * \notxt$. 
Also extra wide documents, i.e., if $w/h\,{>}\,2$, yield a further decrease by 
$\frac{n_{s}}{4} * (\frac{w}{h} - 1)$.
Eventually, the document is scaled to the width (keeping the aspect ration constant) given by 
the $i$-th 
scale, where 
$i\,{=}\,\min(\max(\lfloor i_s\rfloor,0),n_s)$.

%%%%%%%%%%%%%%%%%%%%%%%%%%%%%%%%%%%%%%%%%%%%%%%%%%%%%%%%%%%%%%%%%%%%%%%%%%%%%%%%%%%%
\section{Baseline Post-Processing}
\label{sec:bpp}
%%%%%%%%%%%%%%%%%%%%%%%%%%%%%%%%%%%%%%%%%%%%%%%%%%%%%%%%%%%%%%%%%%%%%%%%%%%%%%%%%%%%

The purpose of post-processing baseline candidates is twofold. First it aims at removing erroneous candidate line segments, and second, it is geared towards combining baseline candidates
horizontally 
to a single baseline, e.g., a polygon representing an entire line in a paragraph.

\smallskip
\paragraph{Error Pruning}
We consider short candidate line segments that either are disoriented, or that are covered by a longer line segment 
as unintended. 
More specifically, let $l_{\mathit{max}}$ be an upper bound for line segments to be considered short, let $\alpha_{\mathit{max}}$ be an orientation threshold, and $d_{\mathit{max}}$ be a distance threshold. Then,  we consider a candidate line segment $l$ to be \emph{short} iff $\nnorm{l}\,{\leq}\,\min(0.2*\bar{l}, l_{\mathit{max}})$, where
$\bar{l}$ is the average length for a set of candidate line segments. 
In a first step we remove short candidate line segments that deviate more than $\alpha_{\mathit{max}}$ degrees from the horizontal. In a second step, we remove short candidate lines that are covered by a longer candidate line segment, where coverage of a line segment 
$l_1\,{=}\,(s_1,e_1)$ by another line segment 
$l_2\,{=}\,(s_2,e_2)$ 
requires two conditions to be satisfied: The projections of $s_1$ and $e_1$ onto the line corresponding to $l_2$ result in points on the segment $l_2$, and the minimal distance of any point on $l_1$ from the line corresponding to $l_2$ is smaller than 
$d_{\mathit{max}}$.

\smallskip
\paragraph{Joining Baseline Segments}
is parametrized by a horizontal ($d_x$), a vertical ($d_y$), and an angular ($d_\alpha$) threshold and proceeds 
recursively 
as follows. 
In a first run, we only allow joins to the right, i.e., the joined baseline segments must not overlap 
horizontally, whereas in a second pass we also allow for overlapping (leftward) joins. 
In particular, for a baseline candidate 
$l_1$, 
another baseline candidate 
$l_2$ 
is a potential join line candidate wrt.~$l_1$ 
iff ($i$) the line containing $s_{2}$ and $e_{2}$ 
does not deviate more than $d_\alpha$ degrees from the horizontal, 
($ii$) $|x_{s_{2}}- x_{e_{1}}|\,{\leq}\, d_x$, 
and ($iii$) $|y_{s_{2}}- y_{e_1}| \,{\leq}\, d_y$. 
In case of non-overlapping joins, we additionally require that 
$|x_{s_{2}}- x_{e_1}|\,{\geq}\, 0$, 
for leftward joins 
$x_{e_{2}}\,{>}\, x_{e_1}$ and $\nnorm{s_{2}-e_1}\,{>}\, d_x/3$ 
must hold in addition. 
Regarding preference, non-overlapping joins are always preferred over leftward joins, 
among non-overlapping joins minimal vertical distance 
prevails, 
among leftward joins it is minimal distance from the end point of 
$l_1$. 
Having determined preferred join line candidates (if any) for every baseline candidate, 
we assemble new candidate baselines (polygons) applying the respective join(s) recursively, starting from any 
base line candidate that itself is not a preferred join line candidate for another baseline.

%%%%%%%%%%%%%%%%%%%%%%%%%%%%%%%%%%%%%%%%%%%%%%%%%%%%%%%%%%%%%%%%%%%%%%%%%%%%%%%%%%%%
\section{Experiments}
\label{sec:exp}
%%%%%%%%%%%%%%%%%%%%%%%%%%%%%%%%%%%%%%%%%%%%%%%%%%%%%%%%%%%%%%%%%%%%%%%%%%%%%%%%%%%%

In this section we first evaluate components and properties of our approach for baseline detection 
individually by means of the cBAD competition\footnote{\url{https://scriptnet.iit.demokritos.gr/competitions/5/}} data and then compare to results on the HisDoc competition\footnote{\url{http://diuf.unifr.ch/main/hisdoc/icdar2017-hisdoc-layout-comp}}. 

\paragraph{The cBAD Dataset and Competition}
was organized into two tracks 
using a dataset that contains various page layouts and degradations.
The data~\cite{DBLP:journals/corr/GruningLDKF17} is composed of historical documents from 9 different archives, the documents span different time periods and 
were split into two sets. 
The first set (\trt{A}) contains 755 samples that are simpler than the 1280 samples in the second set (\trt{B}), since the former just contain text in paragraph form and, moreover, a segmentation into relevant text regions 
is provided. In contrast, \trt{B} comprises more complex samples with various image distortions (e.g.,  
noise, text-lines rotated up to 180\degree),
tables, empty pages, marginalia, and without text region information. 

A subset of the data can be used for training (215, resp.~269, samples) with publicly available annotated baselines serving as groundtruth. In addition, an evaluation tool for baselines is provided that computes precision and recall wrt.~groundtruth lines (cf.~\cite{DBLP:journals/corr/GruningLDKF17} for details), as well as their harmonic mean (F-score). 
The remaining data served as test data with non-public groundtruth used for the competition with F-score as the decisive objective. Our approach yielded a final F-score of 97.13\% on \trt{A}, and without final joining of baselines 85.86\% on \trt{B}, 
winning both tracks~\cite{DBLP:journals/corr/Diem}. 

\begin{figure*}[ht!]
    \centering
    \subfloat[Groundtruth]{
        \includegraphics[width=0.22\textwidth]{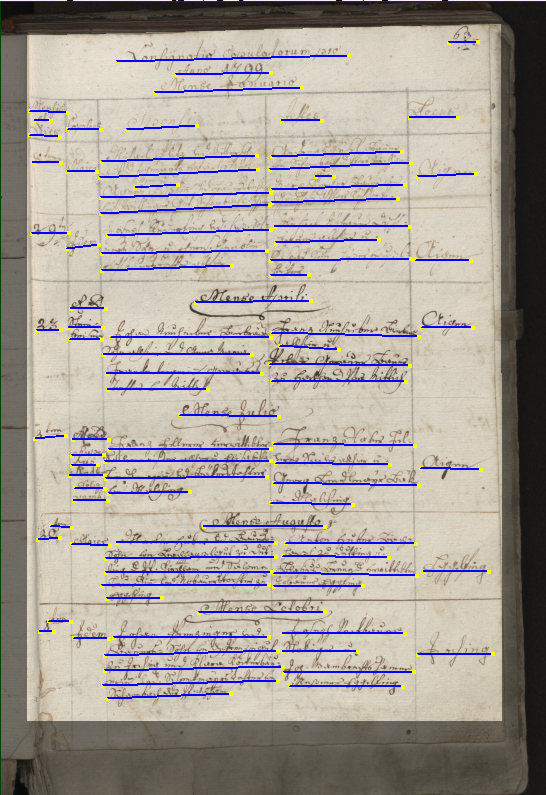}
        \label{fig:mask1gt}
    }
    ~ 
    \subfloat[Detected]{
        \includegraphics[width=0.22\textwidth]{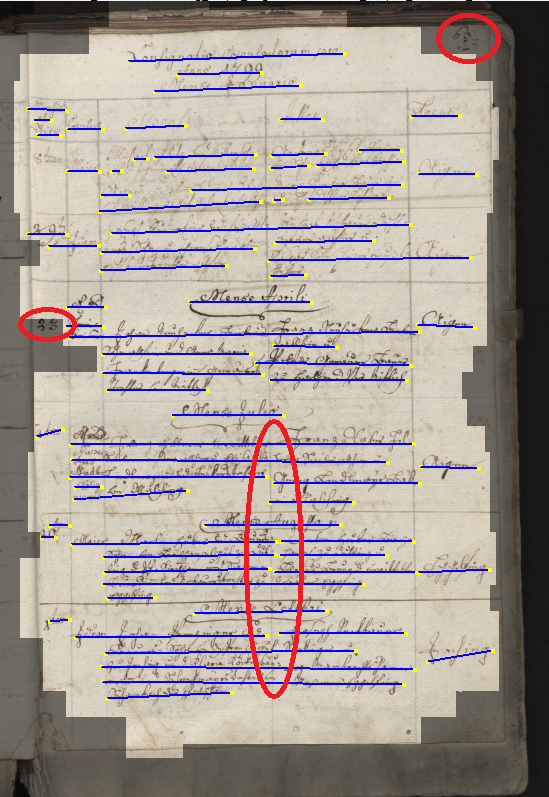}
        \label{fig:mask1dect}
    }
    ~ 
    \subfloat[Groundtruth]{
        \includegraphics[width=0.22\textwidth]{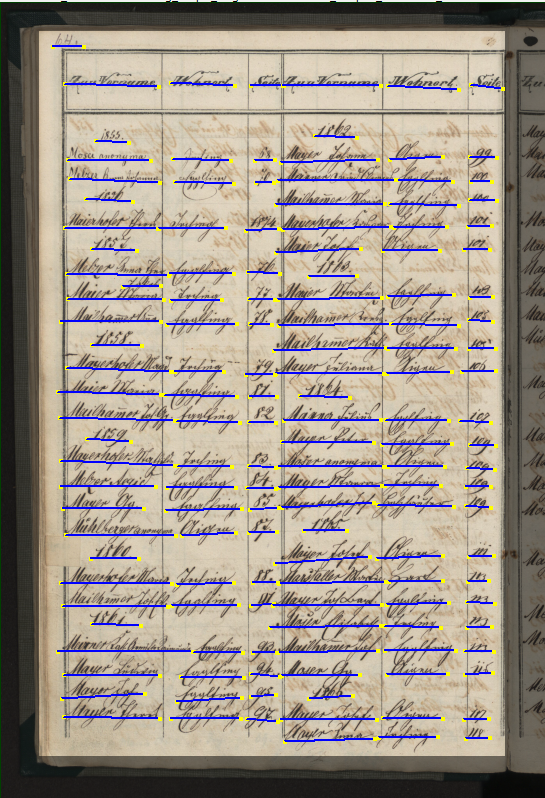}
        \label{fig:mask2gt}
    }
    ~ 
    \subfloat[Detected]{
        \includegraphics[width=0.22\textwidth]{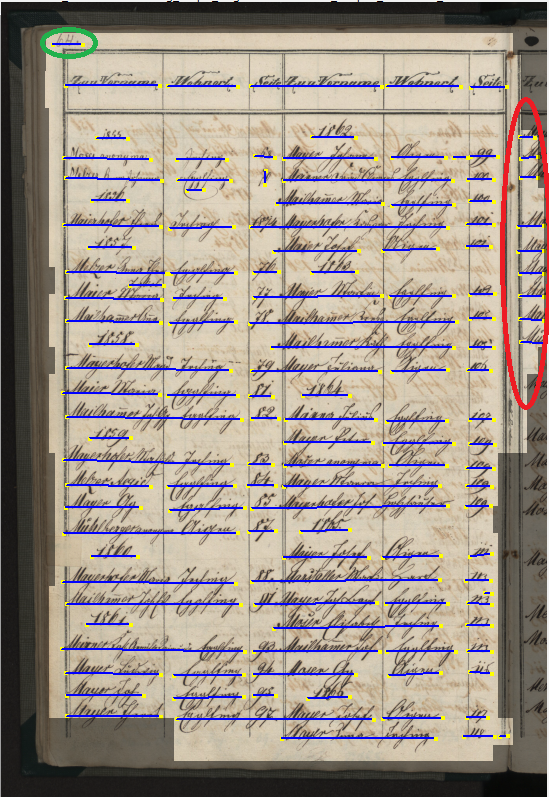}
        \label{fig:mask2dect}
    }
    \caption{Samples of detected regions of interest (including detected baselines) for (the more complex) \trt{B}.}\label{fig:mask}
\vspace*{-3ex}
\end{figure*}

\paragraph{Training and Experimental Setup}
We reserved about 8\% of the training data for cross validation (CVS set:~17, resp.~19, samples). 
Groundtruth binary masks were obtained from the text regions supplied with \trt{A}\ for the \danet, respectively from the baselines provided with both tracks for the \blnet~(turning them into areas using~$5$ pixels height for every line segment).
We additionally labelled the training and CVS data 
wrt.~auxiliary document properties, i.e., whether or not the document is double-page, in landscape format or empty, and regarding leading (quantified into small, normal or large). 

Our nets were trained by Adadelta (decay rate $\rho\,{=}\, 0.95$)  
with a batch size of $2\times 32$ samples. Each sample is a random clipping ($160\times 160$ pixels, resp.~$320\times 320$ pixels) from a prescaled training document image, where the scale is fixed to 192 pixel width for the \danet, and computed individually 
(cf.~Section~\ref{sec:DocumentScaling}) 
for the \blnet. 
We also applied augmentation to 90\% of the samples consisting of random rotation up to $180\degree$ and random scaling by~$\pm 0.5$. Moreover, for the \blnet~a dice coefficient with inner mask $M^I{=}\,M(40,120)$ and $M^Y{=}\,M(37,123)$ has been used.
The post-processing step is also parameterized individually taking into account detected document properties. In particular, given an input 
image of width $w$ 
pixel, we used $l_{\mathit{max}}\,{=}\,w * 0.1$ (resp.~$l_{\mathit{max}}\,{=}\,w * 0.05$ if $\dblp\,{>}\, 0.7$), $\alpha_{\mathit{max}}\,{=}\,30$, and $d_{\mathit{max}}\,{=}\, 20+ 50*\spac$ for error pruning. For joining we set $d_x\,{=}\,w * 0.2$ (resp.~$d_x\,{=}\,w * 0.1$ if $\dblp\,{>}\, 0.7$), $d_y\,{=}\, 20+ 50*\spac$, and $d_\alpha\,{=}\,50$.
\vspace*{-2ex}

\subsection{Component Assessment and cBAD Results}

\paragraph{Simple Document Analysis}
To evaluate the extraction of relevant text regions by the \danet, the dice coefficient of the detected mask wrt.~the groundtruth region has been calculated. Table~\ref{tab:document_analysis} lists the results obtained on the CVS set, as well as for the test data of \trt{A}, since relevant regions are provided for this track. In both cases roughly 90\% of the releveant areas are correctly detected.
Visual inspection reveals that in many cases non-overlapping areas do not contain relevant text and 
result from 
a more fine-grained sampling of the mask compared to the rectangular groundtruth 
(see Fig.~\ref{fig:mask}). However, sometimes also relevant text is missed, e.g., in Fig.~\ref{fig:mask1dect}: 
a table entry and the page number (marked by red ellipses to the left and the upper right), or 
the mask partially contains marginal text as in Fig.~\ref{fig:mask2dect}: 
the page number is contained (green) 
but also spurious text from the next page (red).

\begin{table}[!h]
\centering
\begin{tabular}{c|ccccc}
 & region & landscape & dbl.-page & no text & leading \\
\hline
CVS & 90.89 & 100 & 94.74 & 97.37 & 81.58 \\
\hline
Test &  90.21 & 99.29 & 95.93 & 94.19 & 73.85 \\
\end{tabular}
\caption{Evaluation of simple document analysis (\% GT).}
\label{tab:document_analysis}
\vspace*{-5ex}
\end{table}

Concerning auxiliary document properties, 
outputs for landscape, double-page, and no text are binarized thresholding at $0.5$, and 
leading has been quantified into small, medium, or large using thresholds $0.3$ and $0.6$. 
For post-competition evaluation 
we also labeled the test data wrt.~these properties. While stable results are obtained for the binary properties, 
correctness drops for leading. A more rigorous and consistent labeling than subjective visual classification 
may help to improve in this regard.

\smallskip
\paragraph{Baseline Detection} 
Table~\ref{tab:baseline_detection} reports precision and recall for baselines detected in various settings on the cBAD dataset. Results are listed for both tracks on the CVS set as well as on the test set. 
Entries in the first column correspond to detections obtained by applying the \blnet\ standalone, i.e., with fixed pre-scaling to 1024 pixel width and without  
post-processing (for the test data 
with a precursor of the final \blnet). Values in the second column are obtained applying individual document scaling on behalf of the document properties returned by the \danet, but still without post-processing. Document scaling thus clearly increases precision for both tracks. Moreover, note that the test set for \trt{B}\ does not specify relevant text regions, corresponding results are obtained applying the region masks extracted by the \danet.
  
\begin{table}[!h]
 \centering
\begin{tabular}{c|cccc}
 & \blnet  & no post-p. & add.~pruning & add.~joining\\
 & fixed scale & ind.~scale  &  ind.~scale  &  ind.~scale\\
\hline
CVS A & \!90.88/95.95\! & \!92.73/95.47\! & \!93.28/94.93\! & \!96.92/95.07\!\\
CVS B & \!85.44/85.63\! & \!86.22/88.02\! & \!86.79/87.78\! & n.a.  \\
\hline
Test A & \!76.59/96.76\! & \!81.41/96.27\! & \!91.08/94.78\! & \!{\bf 97.27/96.99}\!\\
Test B & \!76.87/90.08\! & \!78.82/89.04\! & \!{\bf 85.42/86.30}\!  & n.a. \\
\end{tabular}
\caption{Evaluation of baseline detection (\% Prec./\% Rec.).}
\label{tab:baseline_detection}
\vspace*{-5ex}
\end{table}

\begin{figure}[t!]
    \centering
    \subfloat[pruning and joining ok]{
        \includegraphics[width=0.3\textwidth,height=1.5cm]{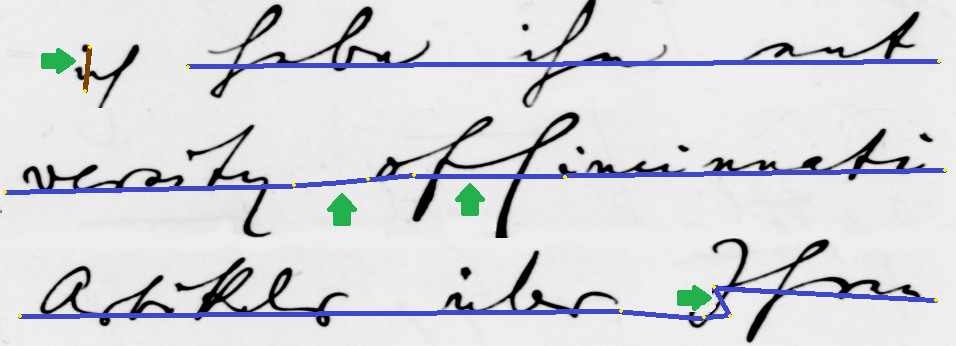}
        \label{fig:postproc1}
    }
    ~ 
    \subfloat[false pruning]{
        \includegraphics[width=0.1\textwidth,height=1.5cm]{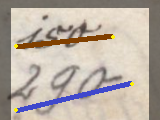}
        \label{fig:postproc2}
    }
    \\[0.5ex]
    \subfloat[unintended join]{
        \includegraphics[width=0.43\textwidth]{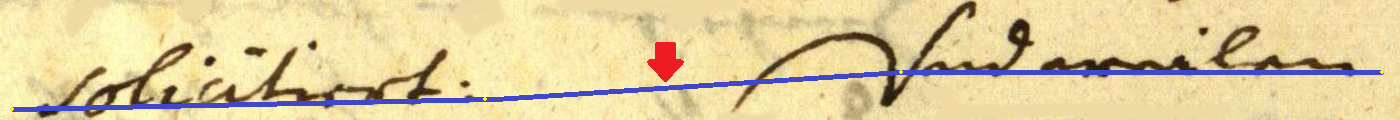}        
        \label{fig:postproc3}
    }
    \caption{Samples of final baselines after post-processing.}\label{fig:postproc}
\vspace*{-3ex}
\end{figure}

\paragraph{Baseline Post-Processing}
Precision increases further when error pruning is applied (cf.~third column in Table~\ref{tab:baseline_detection}). However, besides pruning erroneous lines like the short and almost vertical line in Fig.~\ref{fig:postproc1} (top left green arrow), false positives also eliminate intended lines like in Fig.~\ref{fig:postproc2} (brown line) thus decreasing recall. In summary, it still has a slightly positive effect (increases F-score).

Baseline joining only applies to \trt{A} which is restricted to text in paragraph form. There, however, it achieves a significant improvement (forth column in Table~\ref{tab:baseline_detection}) with an overall increase wrt.~both, precision and recall. 
Unintended joins as in Fig.~\ref{fig:postproc3} are rare (see Fig.~\ref{fig:postproc1} for several positive joins indicated by green arrows, including recursive joins and a leftward join).

\subsection{HisDoc Baseline Detection Results}

For the HisDoc competition 
a collection of pages from three 
medieval manuscripts has been selected with regard to the complexity of their 
layout~\cite{DBLP:journals/corr/Simistra}. 
Compared to 
cBAD the most distinctive characteristic is that they also contain interlinear glosses (explanatory notes) in addition to marginal annotations and additions. The competition comprised three different tasks, one 
(Task~2) addressing baseline extraction. It was laid out to make the competition results comparable to other competitions such as cBAD, e.g., using the same evaluation tool (with parameters: {\small {\tt -tTF 0.75} {\tt -minT 20} {\tt -maxT 20}}).

We thus compare our approach 
using the same parametrization as before except for three minor adaptations: the upper bound on 
document width is changed to reflect the actual 
scale of the 
data ($w_{\mathit{max}}\,{=}\,5000$); the distance threshold for error pruning is enlarged to account for larger leadings due to interline glosses ($d_{\mathit{max}}\,{=}\, 20+ 500*\spac$), while the corresponding 
orientation threshold is decreased ($\alpha_{\mathit{max}}\,{=}\,3$) for stricter short line removal (no tabular data).

\begin{table}[!h]
 \centering
\begin{tabular}{c|cccc}
 & CB55 & CSG18 & CSG863 & Total   \\
\hline
HisDoc best &  {\bf 98.96}  & {\bf 98.79}  & {\bf 98.30}  & {\bf 98.22} \\
\hline
unmodified & 95.73 & 80.79 & 80.46 & 86.01 \\
\blnet~train ctd. & 99.91 & 97.43 & 97.96 & 98.44 \\
\blnet~train new &  {\bf 99.91}  & {\bf 99.25}  & {\bf 98.52}  & {\bf 99.23} \\
\end{tabular}
\caption{Results on HisDoc private test set (\% F-Score).}
\label{tab:hisdoc}
\vspace*{-5ex}
\end{table}

Table~\ref{tab:hisdoc} lists F-score percentages obtained for the three manuscripts individually and in total. The first row repeats the best results reported in~\cite{DBLP:journals/corr/Simistra} for each class 
(rather than the winning system), 
while the second line gives the results of our system without any additional training. The scores reflect 
that the \blnet\ had not been trained 
with interline glosses and is penalized for returning baselines for them. Training the model (just the BL-net) for another 30k iterations (as long as for cBAD) on the HisDoc training samples yields a system scoring slightly better than the 
winning entry. Training the \blnet\ from scratch (also 30k iterations) using color images rather than greyscale images (which we used for cBAD) our approach clearly outperforms the HisDoc contestants in all three categories.

%%%%%%%%%%%%%%%%%%%%%%%%%%%%%%%%%%%%%%%%%%%%%%%%%%%%%%%%%%%%%%%%%%%%%%%%%%%%%%%%%%%%
\section{Discussion and Conclusion}
\label{sec:concl}
%%%%%%%%%%%%%%%%%%%%%%%%%%%%%%%%%%%%%%%%%%%%%%%%%%%%%%%%%%%%%%%%%%%%%%%%%%%%%%%%%%%%

We presented a novel approach to baseline extraction for historical documents that utilizes U-nets and a sliding window approach for simple layout analysis in a pre-processing step, as well as for actual baseline detection. It is complemented by procedural baseline post-processing for pruning spurious candidate lines and potential assembly of candidate line segments into poly-baselines.
It yields convincing results for heterogeneous collections of historical documents exhibiting various degradations (e.g., bleed-through, ornamentations, interlinear glosses), as long as they are of a simple structure (text in paragraph form). To wit, besides winning the 2017 cBAD competition, we showed that it also outreaches the results reported for a corresponding task of the 2017 HisDoc competition.

For more complex structured documents however, there is still potential for further improvement and several issues remain for further research. For instance, increasing the accuracy of relevant text region detection (\danet) would affect recall considerably avoiding the inclusion of marginal text (as in Fig.~\ref{fig:mask2dect}). To this end, dilated residual networks~\cite{DBLP:conf/cvpr/YuKF17} are promising architectures. 

Moreover, extending our simple document analysis towards  multi-dimensional classification in order to segment various document regions (table boundaries, marginalia, ornamentations, etc.) is an obvious next step. E.g., if we had had a more detailed segmentation, we would not have had to retrain the \blnet~in the presence of interlinear glosses (but simply skip baselines from such regions). 
This 
illustrates another benefit of pursuing a modular approach: adaption to different application cases (including glosses or not) not necessarily requires re-training.

Eventually, it seems necessary to consider context for baseline post-processing in complex scenarios. While errors (such as in Figures~\ref{fig:postproc2} and~\ref{fig:postproc3}) were rare in simple settings, preliminary experiments with procedural baseline splitting along vertical lines (in cases as in Fig.~\ref{fig:mask1dect}) indicate a need for more sensitive decisions. 
Resorting to recursive networks, with local spatial context information, 
would be 
an interesting route towards learning corrective 
 post-processing
actions (joining, splitting, pruning).


\nop{%%%NOPed
\section*{Acknowledgment}

Not needed upon submission (unless we obtain feedback before review; no funding organization to acknowledge). Neverteheless kept as a placeholder for incorporating and/or acknowledging anonymous reviewer comments in final version.
}%%%NOPed

%%%%%%%%%%%%%%%%%%%%%%%%%%%%%%%%%%%%%%%%%%%%%%%%%%%%%%%%%%%%%%%%%%%%%%%%%%%%%%%%%%%%

%\makebbltrue

\ifmakebbl 

\bibliographystyle{IEEEtran}
\bibliography{IEEEabrv,paper}

\else

\fi

\end{document}